\documentclass[runningheads]{llncs}
\usepackage[T1]{fontenc}
\usepackage{graphicx}
\usepackage{amsmath}
\usepackage{booktabs}
\usepackage{multirow}
\usepackage[margin=1in]{geometry}
\usepackage{float}
\usepackage[labelfont=bf]{caption}
\usepackage{hyperref}
\usepackage{subcaption}
\hypersetup{
    colorlinks=true,
    linkcolor=blue,
    filecolor=magenta,      
    urlcolor=cyan,
}
\begin{document}
\title{The Critical Role of Model Selection in Causal Inference: A Comparative Analysis of Classification Models within the InferBERT Framework for Pharmacovigilance}
\titlerunning{Model Selection in the InferBERT Framework}
%
\author{Csaba Kiss\inst{1}\orcidID{0009-0005-8523-7481} \and
Roland Molontay\inst{1, 2}\orcidID{0000-0002-0666-5279} \and
Gabriele Pergola\inst{3}\orcidID{0000-0002-7347-2522}}
\authorrunning{C. Kiss et al.}
%
\institute{Department of Stochastics, Institute of Mathematics, Budapest University of Technology and Economics, Műegyetem rkp. 3., Budapest, H-1111, Hungary \and
Institute of Biostatistics and Network Science, Semmelweis University, Üllői út 26., Budapest, H-1085, Hungary \and
Department of Computer Science, University of Warwick, Coventry, CV4 7AL, UK}
\maketitle              
\begin{abstract}
Distinguishing causal adverse drug events (ADEs) from spurious correlations remains a central challenge in pharmacovigilance. The InferBERT framework, which integrates transformer models with Do-calculus, offers a promising solution, but its efficacy hinges on the performance of its underlying classification model.
This study systematically evaluates the impact of the classification model within the InferBERT framework. We assess whether a simpler statistical model are sufficient, if domain-specific pre-training provides an advantage, and if scaling to large language models (LLMs) improves causal signal detection. We also investigate the effect of post-hoc probability calibration.

We conducted a rigorous comparative study on two pharmacovigilance benchmarks: Analgesics-induced Acute Liver Failure (AILF) and Tramadol-related Mortalities (TRAM). We evaluated four models: XGBoost (a baseline), ALBERT (the original InferBERT model), BioBERT (a biomedical transformer), and Med-LLaMA (a medical LLM). Using a 5-fold cross-validation design repeated over 20 runs, we measured predictive accuracy, Expected Calibration Error (ECE) before and after isotonic regression, and the concordance of discovered causal terms with traditional methods (PRR, ROR, EBGM) via the Jaccard index. Statistical significance was assessed with paired t-tests.

BioBERT demonstrated statistically significant superiority in predictive accuracy on both datasets ($p < 0.0001$ in all pairwise comparisons). In contrast, the larger Med-LLaMA model, fine-tuned via parameter-efficient methods, consistently underperformed, ranking last. Domain-specific pre-training was the decisive factor for success. Probability calibration reliably improved ECE but had an inconsistent and sometimes negative effect on accuracy and causal discovery. Critically, BioBERT's predictive superiority translated into the highest concordance with traditional pharmacovigilance signals, indicating more robust causal term identification.
The choice of classification model is a critical determinant of the InferBERT framework's success. Our findings establish that domain-specific pre-training, as embodied by BioBERT, provides a decisive advantage over both simpler models and larger LLMs. For computational pharmacovigilance, investing in domain-aware, but managable sized models is more impactful than scaling model size.

\end{abstract}

\section{Introduction}

Pharmacovigilance is the cornerstone of post-market drug safety, tasked with identifying, evaluating, understanding, and preventing adverse drug events (ADEs) from observational data \cite{alomar2020post,WHO04}. 
To monitor adverse drug events, organizations rely on large-scale spontaneous reporting systems, such as the FDA Adverse Event Reporting System (FAERS) \cite{potter2025fda}. While FAERS provides an unprecedented volume of real-world observational data, it presents a significant computational challenge: extracting true \textit{safety signals} from an overwhelming background of observational \textit{noise} \cite{hernan2010causal,zhao2022machine}. In this context, a safety signal represents a credible, potentially causal link between a drug and an adverse event, whereas noise comprises spurious statistical correlations driven by patient comorbidities, co-prescribed medications, and inherent reporting biases. Modern machine learning models excel at identifying complex statistical correlations within these vast databases, but without causal grounding, they frequently mix this noise with genuine safety signals \cite{hernan2010causal,zhao2022machine}.

To bridge the gap between predictive modeling and formal causal reasoning, recent methodologies have aimed at integrating neural representations with structural causal models. Notably,
the InferBERT framework introduced a novel, two-stage pipeline for processing FAERS data by integrating neural models with formal causal principles \cite{wang2021inferbert}: (i) a neural classification model first predicts a clinical outcome based on patient reports, and subsequentially, (ii) the Judea Pearl's Do-calculus is applied to the model's probabilistic outputs to simulate interventions and identify potential causal factors.    

However, the entire InferBERT pipeline is critically dependent on the quality of the initial classification stage. The accuracy of the predictions and the calibration of the output probabilities serve as the foundation for the subsequent causal analysis. While the original study validated the framework's concept, it does not systematically analyse how sensitive the causal conclusions are to the predictive component. 
In particular, it is unclear whether improved predictive performance reliably translates into improved causal signal discoveries. It is also unclear whether a transformer-based encoder is necessary, or whether a strong tabular baseline such as XGBoost can yield comparable causal outputs. And beyond model class, the contribution of domain knowledge remains uncertain: biomedical pre-training may lead to different causal conclusions than general-purpose representations. It is further unclear whether increasing the scale of models via large medical language models (LLMs) improves causal signal detection. Finally, since the causal procedure relies on probability estimates, it remains an open question the impact of post-hoc calibration on the models confidence and identification of causal relations.

This study addresses these questions through a rigorous, multi-faceted comparative analysis. We hypothesize that a model's predictive accuracy and the quality of its probability estimates are directly linked to its ability to identify clinically relevant causal signals. Our primary contribution is a comprehensive evaluation of four distinct modeling paradigms within the InferBERT framework: (i) XGBoost \cite{chen2016xgboost}, representing a powerful and efficient non-transformer baseline ; (ii) ALBERT \cite{lan2019albert}, to benchmark against the original InferBERT implementation; (iii) BioBERT \cite{lee2020biobert}, a transformer pre-trained on biomedical literature, to test the impact of domain-specific knowledge ; and (iv) Med-LLaMA \cite{xie2024mellama}, a large medical language model, to investigate the utility of scale.
%
Using two public pharmacovigilance data sets, we assess the models' performance along three dimensions: (1) predictive accuracy, (2) probability calibration, and (3) concordance of identified causal terms with established pharmacovigilance methods. The experimental analyses quantify how model choice affects both predictive behaviour and downstream causal discovery, yielding a consistent ranking across settings and providing empirical guidance for selecting the classification component in InferBERT-style causal inference pipelines for pharmacovigilance.

\section{Materials and methods}

\subsection{Causal inference framework}
\begin{figure}[H]
    \centering
        \includegraphics[width=\textwidth]{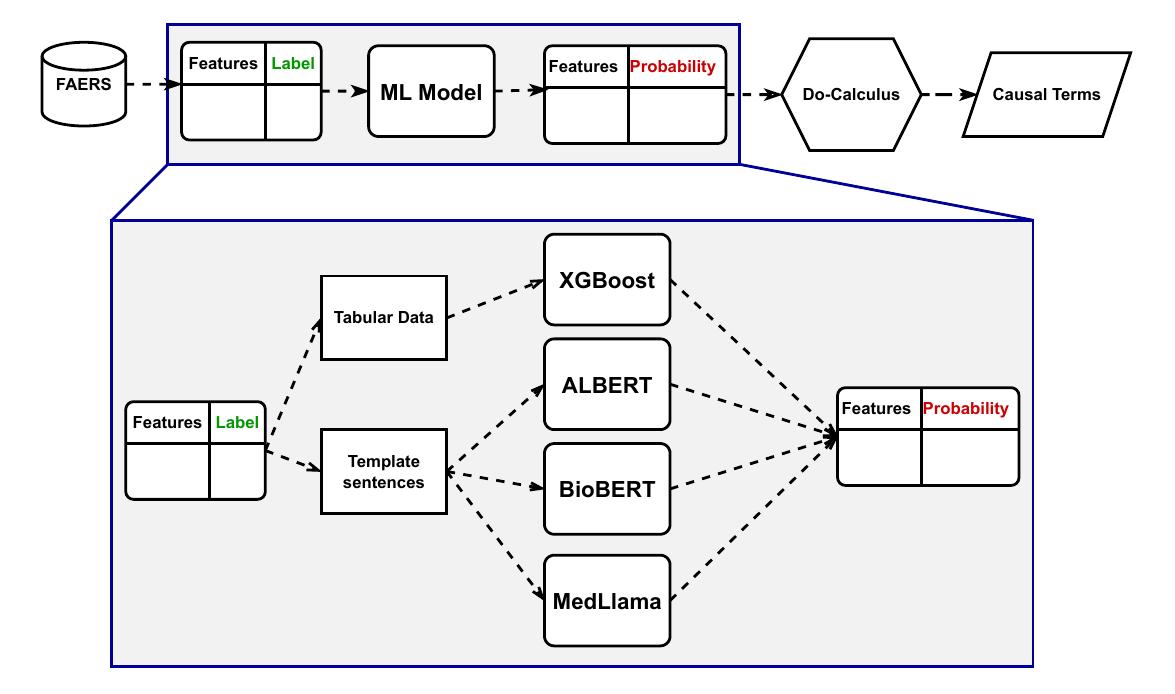}
        \caption{Schematic overview of the framework and the experimental setup. The upper panel depicts the full causal inference framework, flowing from FAERS data extraction to causal term identification via Do-calculus. The lower panel illustrates the specific experimental setup for the different models, contrasting the tabular data processing for XGBoost with the template-based sentence generation used for the transformer models (ALBERT, BioBERT, and Med-LLaMA) as input.}
        \label{fig:flowchart}
\end{figure}

Our methodology adheres to the two-stage InferBERT process, which is visually summarized in the upper panel of Fig. \ref{fig:flowchart}:
\begin{enumerate}
\item \textbf{Probabilistic Classification:} A model is trained to predict the probability of the clinical endpoint. As illustrated in the lower panel of Fig. \ref{fig:flowchart}, we evaluated different experimental setups for this stage: for transformer models (ALBERT, BioBERT, Med-LLaMA), structured features are formatted into a natural language sentence via a fixed template; for XGBoost, features are multi-hot encoded directly from the tabular data.
\item \textbf{Causal Analysis with Do-Calculus:} The model's predicted probabilities are used to perform causal interventions. Following the original InferBERT methodology, a one-tailed z-test identifies statistically significant terms ($p < 0.05$), which are considered causal factors \cite{pearl2009causal}.
\end{enumerate}

\subsection{Experimental design and statistical analysis}
We employed a robust cross-validation scheme to ensure reliable performance estimates. For each dataset, the data was randomly split into five folds. We performed 20 independent runs, each using a unique combination of three folds for training, one for validation (e.g., hyperparameter tuning, early stopping, and calibrator fitting), and one for testing. This ensures every data subset is used for testing exactly four times.

Model performances were compared using paired t-tests on the metrics collected from the 20 test sets. A result was deemed statistically significant if the p-value was less than 0.05. We report the median performance across the 20 runs to mitigate the influence of outliers.

\subsection{Evaluation metrics}
We evaluated the models across three key dimensions that directly map to our research questions.

\noindent\textbf{1. Predictive performance:} Measured using standard classification \textbf{accuracy}. A higher value indicates a better ability to predict the clinical outcome.

\noindent\textbf{2. Probability quality:} Assessed with the \textbf{Expected Calibration Error (ECE)} \cite{posocco2021estimatingexpectedcalibrationerrors}, which quantifies the discrepancy between a model's predicted probabilities (confidence) and the observed frequencies (accuracy). ECE is computed by grouping predictions into $M$ equally-spaced confidence bins and calculating a weighted average of the bin-wise calibration error:
\[
\text{ECE} = \sum_{m=1}^{M} \frac{|B_m|}{n} \left| \text{acc}(B_m) - \text{conf}(B_m) \right|
\]
where $B_m$ is the set of samples in bin $m$, $\text{acc}(B_m)$ is the accuracy of $B_m$, and $\text{conf}(B_m)$ is the average confidence in $B_m$. A lower ECE signifies better-calibrated probabilities. We evaluated ECE on the raw model outputs and after applying \textbf{Isotonic Regression}, a post-processing calibration method trained on the validation set.

\noindent\textbf{3. Causal discovery concordance:} To measure the quality of the identified causal terms, we calculated the \textbf{Jaccard Index}. This index measures the overlap between the set of causal terms identified by our models and those flagged by three traditional pharmacovigilance methods: Proportional Reporting Ratio (PRR), Reporting Odds Ratio (ROR), and Empirical Bayes Geometric Mean (EBGM) \cite{wang2021inferbert}. A higher Jaccard index suggests greater concordance with established signals.

\subsection{Model Implementation and Training}
All transformer models were fine-tuned for a maximum of three epochs with a learning rate $\mu= 2e-5$. To prevent overfitting, we employed an early stopping strategy based on the validation set loss, with a patience of five evaluation steps.

\noindent\textbf{ALBERT and BioBERT} were based on the `textattack/albert-base-v2-imdb' and `dmis-lab/biobert-base-cased-v1.1` models, with approximately 12 million and 110 million parameters, respectively. These models were fully fine-tuned on the downstream classification task. 

\noindent\textbf{Med-LLaMA}, specifically the 8-billion parameter `YBXL/Med-LLaMA3-8B' model \footnote{https://huggingface.co/YBXL/Med-LLaMA3-8B}, was fine-tuned adoping a parameter-efficient fine-tuning (PEFT) approach. In particular, the model was loaded in 4-bit precision using `nf4' quantization. We then applied Low-Rank Adaptation (LoRA) \cite{le2024impact}, using conventional parameters of rank (r) of 8 and alpha of 16, targeting the `q\_proj' and `v\_proj' modules. Training was performed using the memory-efficient `paged\_adamw\_8bit` optimizer. This approach significantly reduces the number of trainable parameters,  while preserving and specilising the performance of the models.

It is worth noticing, that this PEFT setup, while efficient, may limit the model's ability to fully adapt to the task compared to full fine-tuning, a point we revisit in Section \ref{sec:discussion}.

Code and full experimental results (including AUC scores across all cross-validation splits and causal trees) are available at \url{github.com/hsdslab/biomedical-causal-inference.git}.

\subsection{Data and preprocessing}
We utilized two publicly available datasets from the original InferBERT study \cite{wang2021inferbert}:
\begin{enumerate}  
    \item \textbf{Analgesics-induced Acute Liver Failure (AILF):} FAERS reports where the endpoint is the binary occurrence of acute liver failure.
    \item \textbf{Tramadol-related Mortalities (TRAM):} FAERS reports where the endpoint is patient death (binary).
\end{enumerate}  

We expanded the original feature set to include secondary suspect, concomitant, and interacting drugs, as detailed in Table \ref{tab:features}, to provide a more comprehensive representation of a patient's drug exposure. Preprocessing involved standardizing drug doses, binning patient age, removing reports with missing endpoints, and deduplicating drug entries to prevent data leakage.

\begin{table}[H]
\centering
\caption{Features, endpoints, and report counts for the two datasets.}
\label{tab:features}
\begin{tabular}{@{}lll@{}}
\toprule
\textbf{Dataset} & \textbf{Analgesics-induced liver failure} & \textbf{Tramadol related mortalities} \\ \midrule
\multirow{9}{*}{\textbf{Features}} & Age & Age \\
& Dose & Dose \\
& Gender (Male/Female) & Gender (Male/Female) \\
& Primary Suspect Drugs (psd) & Primary Suspect Drugs (psd) \\
& Secondary Suspect Drugs (ssd) & Secondary Suspect Drugs (ssd) \\
& Concomitant Drugs (ccd) & Concomitant Drugs (ccd) \\
& Interacting Drugs (idrug) & Interacting Drugs (idrug) \\
& Indication & Adverse Drug Event (ade) \\
& Outcome & \\ \midrule
\textbf{Endpoint} & Adverse Drug Event (ade) & Outcome \\
\textbf{(Label)} & \textit{1: acute liver failure, 0: otherwise} & \textit{1: death, 0: otherwise} \\ \midrule
\textbf{Total Reports} & 36,661 & 27,245 \\ \bottomrule
\end{tabular}
\end{table}

\section{Results}

Our results systematically address the core questions of model selection, scale, and calibration within the InferBERT framework. We report the median performance of each model across the 20 cross-validation runs.

\subsection{Superior Predictive Accuracy of the Domain-Specific Model}

\begin{figure}[H]
    \centering
    \begin{subfigure}{0.49\textwidth}
        \includegraphics[width=\linewidth]{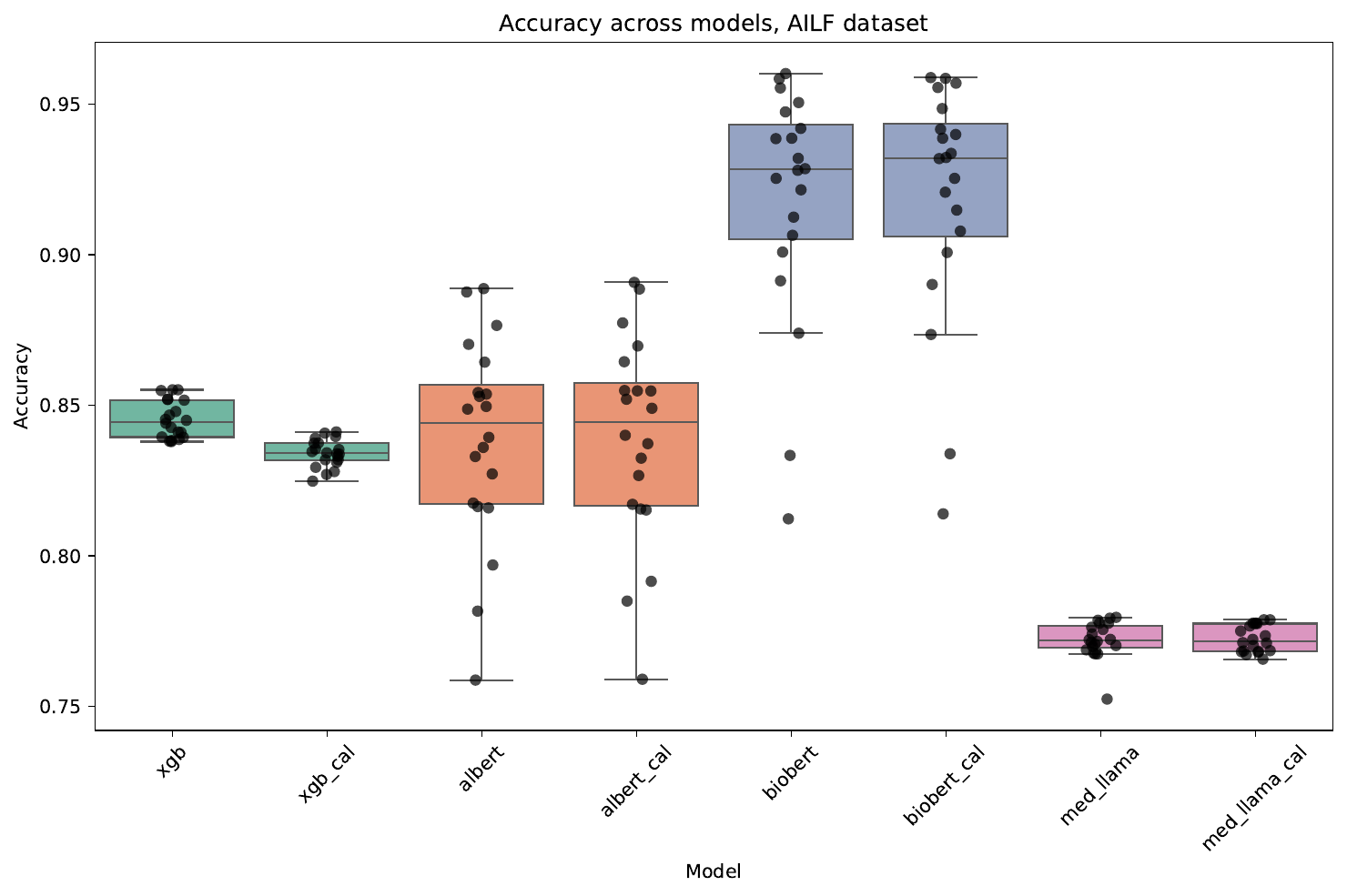}
        \caption{Accuracy on AILF Dataset}
        \label{fig:acc_ailf}
    \end{subfigure}
    \hfill
    \begin{subfigure}{0.49\textwidth}
        \includegraphics[width=\linewidth]{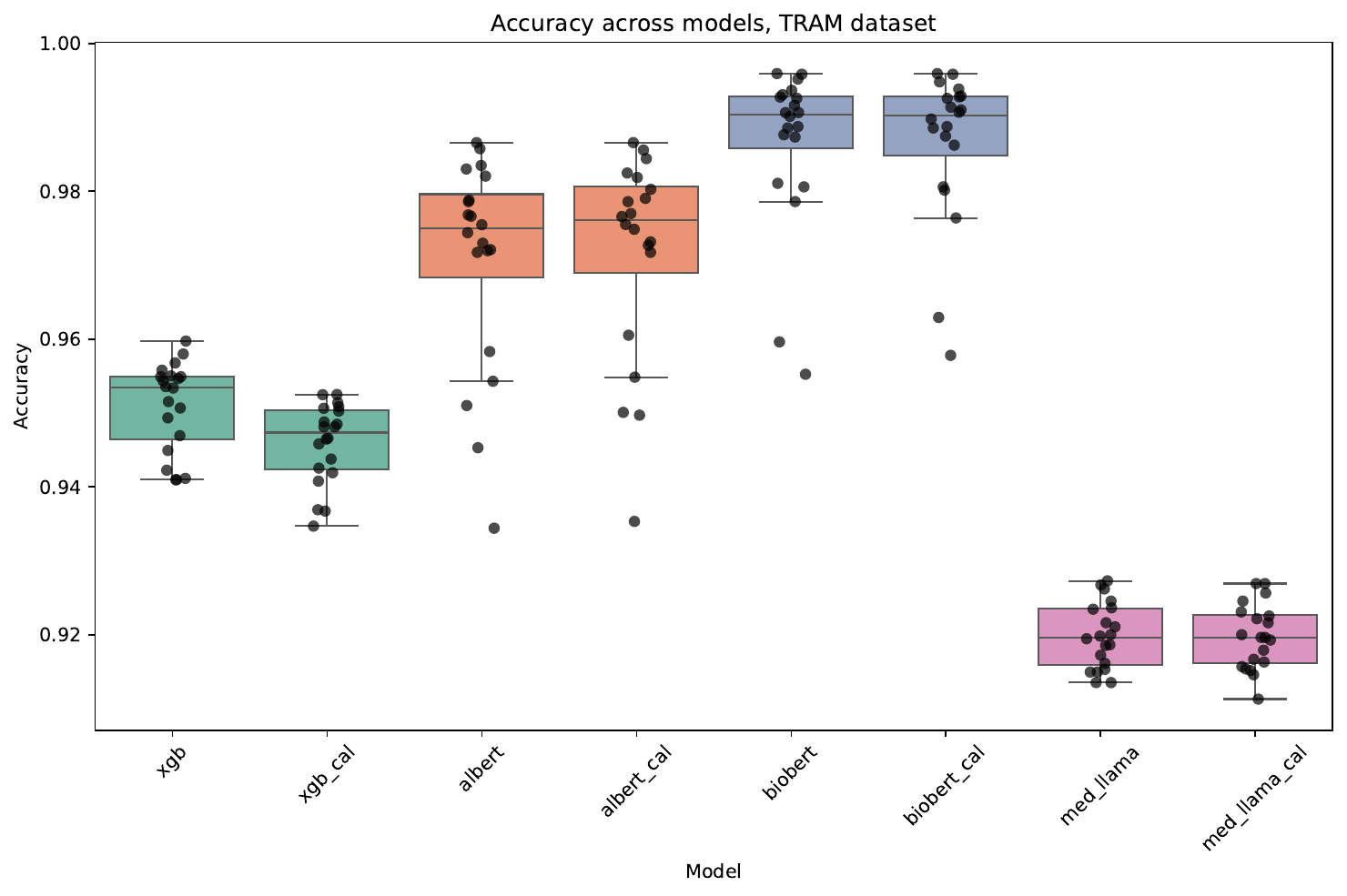}
        \caption{Accuracy on TRAM Dataset}
        \label{fig:acc_tram}
    \end{subfigure}
    \caption{Distribution of classification accuracy across 20 runs for each model. The central mark indicates the median, the box extends to the 25th and 75th percentiles, and the whiskers show the range of non-outlier data. BioBERT demonstrates consistently higher accuracy and lower variance.}
    \label{fig:acc_plots}
\end{figure}

First, we evaluated the models' fundamental ability to predict clinical outcomes. As shown in Table \ref{tab:accuracy_results} and Figure \ref{fig:acc_plots}, \textbf{BioBERT achieved the highest accuracy on both datasets, and the improvements were statistically significant against all other models.} This establishes domain-specific pre-training as the key differentiator for predictive performance.

On the AILF dataset, BioBERT's median accuracy (0.928) was substantially higher than that of ALBERT (0.844) and XGBoost (0.845). The paired t-tests confirmed this dominance, showing BioBERT to be significantly better than both ($p < 0.0001$). Conversely, the large-scale Med-LLaMA model was the poorest performer (0.772), significantly worse than even the XGBoost baseline ($p < 0.0001$).

This trend was replicated on the TRAM dataset. BioBERT (0.990) again significantly outperformed ALBERT (0.975, $p < 0.0001$) and XGBoost (0.954, $p < 0.0001$). These results provide a clear answer to our first two research questions: while the original InferBERT (ALBERT) offers an advantage over the XGBoost baseline on the TRAM dataset, the most significant performance gain comes from leveraging a domain-specific model like BioBERT. The hypothesis that a larger LLM would perform better was not supported; indeed, scale was detrimental in this context.

\begin{table}[H]
\centering
\caption{Median classification accuracy across 20 cross-validation runs. Higher is better. The best performing model in each column is in bold. Paired t-tests confirm BioBERT's superiority over all other models on both datasets ($p < 0.0001$).}
\label{tab:accuracy_results}
\begin{tabular}{@{}lcc@{}}
\toprule
\textbf{Model} & \textbf{AILF Accuracy} & \textbf{TRAM Accuracy} \\ \midrule
XGBoost & 0.8445 & 0.9535 \\
XGBoost Calibrated & 0.8341 & 0.9474 \\ \midrule
ALBERT & 0.8441 & 0.9749 \\
ALBERT Calibrated & 0.8445 & 0.9760 \\ \midrule
BioBERT & 0.9283 & \textbf{0.9904} \\
BioBERT Calibrated & \textbf{0.9321} & 0.9902 \\ \midrule
Med-LLaMA & 0.7719 & 0.9196 \\
Med-LLaMA Calibrated & 0.7717 & 0.9196 \\ \bottomrule
\end{tabular}
\end{table}

\subsection{The Nuanced and Often Limited Impact of Probability Calibration}
Next, we investigated whether post-hoc calibration reliably improves the framework. As expected, applying isotonic regression consistently improved the ECE for most models (Table \ref{tab:ece_results}), indicating that the calibrated probabilities were more reliable. For example, calibration significantly reduced ALBERT's ECE on the TRAM dataset from 0.0137 to 0.0071 ($p < 0.0001$).

However, this improvement in probability quality came with a frequent and significant accuracy trade-off. Calibration significantly degraded XGBoost's accuracy on both datasets ($p < 0.0001$). For the transformer models, the effect was mixed: it provided a small but significant boost to BioBERT on AILF ($p = 0.0215$) and ALBERT on TRAM ($p = 0.0049$), but otherwise had no significant effect. This confirms that calibration is not a universally beneficial operation and its impact on predictive accuracy is model-dependent. Its value for the downstream causal task must therefore be carefully evaluated.

\begin{table}[H]
\centering
\caption{Median Expected Calibration Error (ECE) across 20 runs. Lower is better. The best performing model in each column is in bold.}
\label{tab:ece_results}
\begin{tabular}{@{}lcc@{}}
\toprule
\textbf{Model} & \textbf{AILF ECE} & \textbf{TRAM ECE} \\ \midrule
XGBoost & 0.0203 & 0.0125 \\
XGBoost Calibrated & \textbf{0.0120} & 0.0095 \\ \midrule
ALBERT & 0.0308 & 0.0137 \\
ALBERT Calibrated & 0.0245 & 0.0071 \\ \midrule
BioBERT & 0.0263 & 0.0068 \\
BioBERT Calibrated & 0.0168 & \textbf{0.0040} \\ \midrule
Med-LLaMA & 0.0178 & 0.0133 \\
Med-LLaMA Calibrated & 0.0150 & 0.0115 \\ \bottomrule
\end{tabular}
\end{table}

\subsection{Superior Predictive Accuracy Drives Robust Causal Discovery}
Finally, we assessed how model choice and calibration affect the ultimate output of the framework: the set of identified causal terms. Table \ref{tab:jaccard_results} shows the Jaccard index comparing the models' outputs to traditional methods.

\textbf{Critically, BioBERT, the most accurate model, also demonstrated the strongest concordance with PRR and EBGM signals, establishing a direct link between predictive performance and causal utility.} On the AILF dataset, calibrated BioBERT achieved the highest Jaccard index vs. PRR (0.0517) and EBGM (0.0085). On the TRAM dataset, it was also the top performer against PRR (0.3166) and EBGM (0.0761). This strong alignment suggests that a model with superior predictive capabilities provides a more reliable foundation for the Do-calculus, yielding signals that align with established pharmacovigilance heuristics. While XGBoost showed high concordance with the Reporting Odds Ratio (ROR), BioBERT's strong and consistent performance across multiple benchmarks reinforces its position as the most reliable model for causal signal detection.

The effect of calibration on causal discovery was again inconsistent. For some models, better ECE appeared to help. For instance, calibrating ALBERT on the AILF dataset significantly improved its Jaccard index vs. ROR ($p = 0.0282$). In contrast, calibrating XGBoost significantly degraded its concordance vs. PRR ($p = 0.0220$). This suggests that while probability quality is important, the interaction between a model's predictive patterns and the calibration process is complex, and a universal recommendation for its current use cannot be established.

\begin{table}[H]
\centering
\caption{Median Jaccard Index of causal terms vs. traditional methods. Higher is better. The best performing model for each comparison is in bold.}
\label{tab:jaccard_results}
\resizebox{0.88\textwidth}{!}{%
\begin{tabular}{@{}lcccccc@{}}
\toprule
& \multicolumn{3}{c}{\textbf{AILF Dataset}} & \multicolumn{3}{c}{\textbf{TRAM Dataset}} \\
\cmidrule(r){2-4} \cmidrule(l){5-7}
\textbf{Model} & \textbf{vs. PRR} & \textbf{vs. ROR} & \textbf{vs. EBGM} & \textbf{vs. PRR} & \textbf{vs. ROR} & \textbf{vs. EBGM} \\ \midrule
XGBoost & 0.0494 & 0.4665 & 0.0078 & 0.2977 & \textbf{0.7897} & 0.0749 \\
XGBoost Cal. & 0.0476 & 0.4616 & 0.0075 & 0.3069 & 0.7844 & 0.0753 \\ \midrule
ALBERT & 0.0434 & 0.4722 & 0.0074 & 0.3044 & 0.7777 & 0.0757 \\
ALBERT Cal. & 0.0415 & 0.4800 & 0.0073 & 0.3046 & 0.7716 & 0.0753 \\ \midrule
BioBERT & 0.0515 & 0.4296 & \textbf{0.0085} & \textbf{0.3166} & 0.7767 & \textbf{0.0761} \\
BioBERT Cal. & \textbf{0.0517} & 0.4338 & \textbf{0.0085} & 0.3132 & 0.7782 & 0.0757 \\ \midrule
Med-LLaMA & 0.0418 & 0.4579 & 0.0072 & 0.2857 & 0.7447 & 0.0733 \\
Med-LLaMA Cal. & 0.0407 & \textbf{0.4686} & 0.0071 & 0.2842 & 0.7566 & 0.0729 \\ \bottomrule
\end{tabular}
}
\end{table}

\subsection{Qualitative Analysis Reveals Nuanced Causal Structures}
To provide a more qualitative understanding of the causal relationships identified by each model, we constructed causal trees for the top-performing models. Following the methodology proposed by Wang et al. \cite{wang2021inferbert}, these trees are constructed hierarchically. The process begins by identifying the 'root cause'—the term with the highest z-score. Then, a recursive do-calculus analysis is performed on the subset of the data containing the root cause to find secondary and tertiary causal factors, visualizing the potential causal pathways.

\begin{figure}[H]
    \centering
    \begin{subfigure}{\textwidth}
        \centering
        \includegraphics[width=\textwidth]{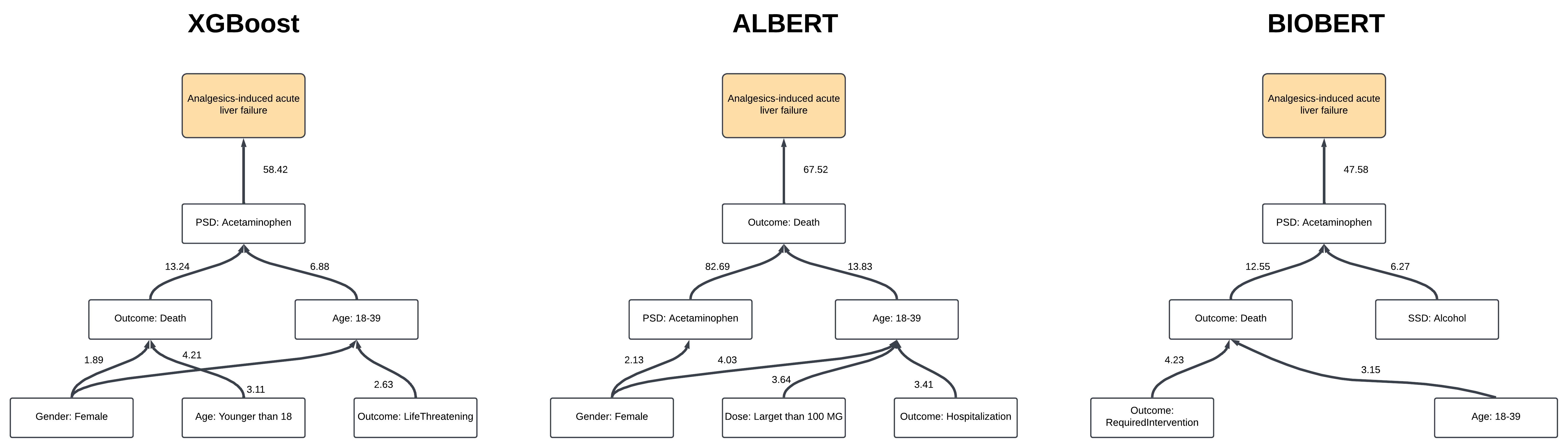}
        \caption{Analgesics-induced Acute Liver Failure (AILF) Dataset}
        \label{fig:causal_trees_ailf}
    \end{subfigure}
    \vspace{1em}
    \begin{subfigure}{\textwidth}
        \centering
        \includegraphics[width=\textwidth]{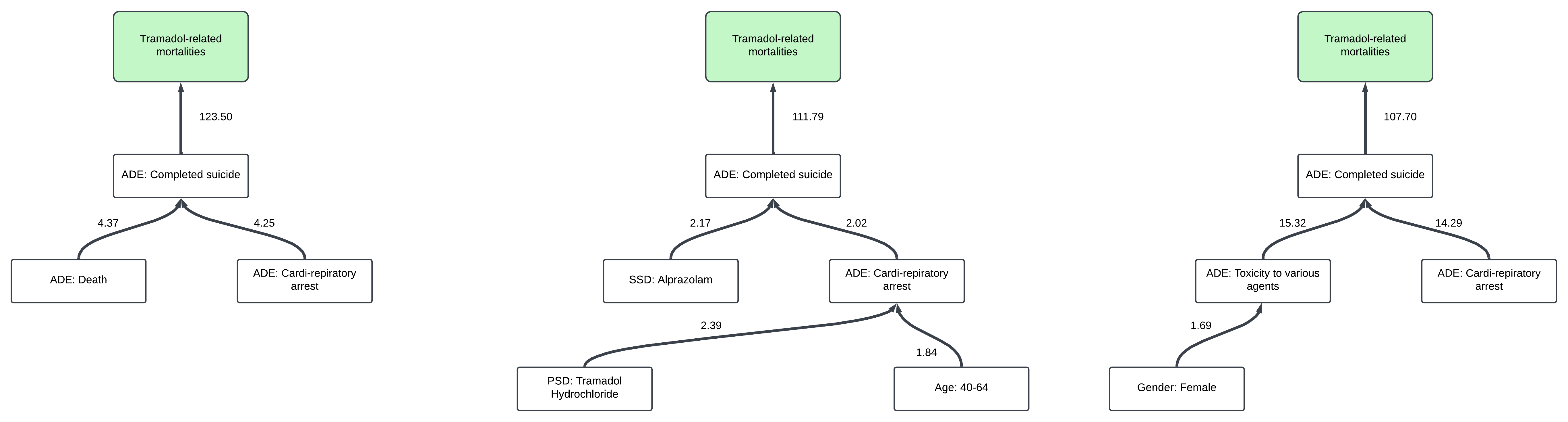}
        \caption{Tramadol-related Mortalities (TRAM) Dataset}
        \label{fig:causal_trees_tram}
    \end{subfigure}
    \caption{Causal trees generated by XGBoost, ALBERT, and BioBERT (from left to right in each panel). These trees are derived from a single, representative cross-validation fold to provide a qualitative illustration of the models' inferred causal hierarchies. Note BioBERT's identification of clinically plausible factors like `alcohol` (A) and `female gender` (B), which are missed by the other models.}
    \label{fig:causal_trees}
\end{figure}

Figure \ref{fig:causal_trees} illustrates these causal trees, derived from a single, representative cross-validation fold. A visual inspection reveals a general alignment in the primary causal factors identified across the models. However, notable differences emerge, particularly with the more accurate BioBERT model, suggesting it captures more nuanced and clinically plausible causal signals.

For the AILF dataset (Figure \ref{fig:causal_trees}A), all models identify `PSD: acetaminophen` as the primary or secondary root cause. Uniquely, however, BioBERT identifies `ssd: alcohol` as a significant secondary factor that the other models failed to detect. Similarly, in the TRAM dataset (Figure \ref{fig:causal_trees}B), BioBERT's tree includes `gender: female` as a causal factor. Conversely, ALBERT identifies the secondary suspect drug (`ssd`) `alprazolam` as a factor, a term that does not appear in BioBERT's hierarchy. These distinctions suggest that the superior predictive accuracy and domain knowledge of BioBERT may allow it to surface different and potentially more clinically relevant causal hypotheses.

\section{Discussion}
\label{sec:discussion}
Our systematic evaluation provides a clear and actionable hierarchy of model performance within the InferBERT framework, with the most definitive finding being the decisive advantage of domain-specific pre-training. The superior performance of BioBERT, which significantly outperformed all other models in predictive accuracy, underscores the profound value of embedding domain knowledge directly into the model's parameters \cite{gu2021domain}. While BioBERT ($\sim$110M parameters) is larger than the general-purpose ALBERT model ($\sim$12M parameters), its key advantage appears to stem from its training on biomedical literature (PubMed, PMC). This equips it with an intrinsic, pre-existing understanding of medical terminology, pathophysiology, and drug-effect relationships, which translates directly into a more powerful and accurate model for this clinical prediction task.

In contrast to BioBERT, Med-LLaMA (8B parameters) was the weakest performer in our experiments. This provides a useful counterpoint to the common assumption that larger models will automatically improve downstream performance, at least for structured prediction settings derived from tabular features \cite{huertas2024gradient}. The result is likely explained by a combination of factors. First, the parameter-efficient fine-tuning (PEFT) setup, while widely adopted and computationally efficient, may constrain how fully the model can adapt to this specific task relative to full fine-tuning of a smaller model \cite{le2024impact}. Second, the input representation may place the model at a disadvantage: converting structured variables into a single templated sentence can obscure relational information that tabular methods such as XGBoost exploit directly, and it may not align well with the strengths of LLMs trained primarily for open-ended language modelling and generation.

Crucially, these differences in predictive performance directly impacted the primary goal of the framework: causal discovery. Our results strongly support the hypothesis that a more accurate predictive model serves as a better foundation for causal inference. A model with higher accuracy has likely learned a more faithful representation of the complex relationships within the data-generating process. This more accurate "world-model" in turn, allows the Do-calculus interventions to yield more meaningful and reliable results. BioBERT's superior accuracy therefore translated into a higher concordance with signals identified by established pharmacovigilance methods like PRR and EBGM, establishing a clear link between predictive power and causal utility. The qualitative analysis of causal trees further strengthens this, showing that BioBERT identified clinically plausible risk factors (e.g., alcohol, female gender) that were missed by other models.

The role of post-hoc probability calibration, however, was more nuanced. While isotonic regression consistently improved the Expected Calibration Error (ECE), its overall benefit to the framework was ambiguous. Its impact on predictive accuracy was inconsistent, and its effect on the final set of causal terms was highly model-dependent, improving concordance for some models while degrading it for others. This indicates that calibration should not be treated as a mandatory post-processing step. Instead, calibration should be treated as an optional step and evaluated in terms of its impact on the downstream causal analysis, since it can alter the probability estimates on which the Do-calculus procedure relies.

\subsection{Limitations and Future Work}
This study has certain limitations that suggest avenues for future research. First, our use of traditional signal detection methods as a benchmark serves as a pragmatic measure for alignment with current epidemiological standards, but it does not represent a definitive causal ground truth. Validation against expert clinical review or results from randomized controlled trials would be invaluable for confirming the clinical relevance of the identified causal terms. Second, the templated conversion of structured data into a single sentence is a significant bottleneck. This approach forces an unnatural, prose-based representation onto inherently structured features and likely fails to capitalize on the relational inductive biases that specialized architectures can offer. Future work should therefore investigate multimodal or multi-input architectures—models that can process structured features (e.g., drug lists, demographics) and any available free-text narratives in parallel through distinct, modality-specific encoders before integrating them for prediction \cite{wang2022transtab}. Such an approach would respect the native format of each data type and could unlock more nuanced predictive and causal signals. Finally, exploring more powerful fine-tuning techniques for LLMs on structured data, or developing more suitable input representations, could help determine if their scale can be effectively harnessed for this class of problems.

\section{Conclusion}

In this comprehensive comparative study, we have demonstrated that the performance of the InferBERT causal inference framework is critically dependent on its underlying classification model. A domain-specific transformer, BioBERT, decisively outperforms a traditional machine learning baseline, a general-purpose transformer, and a large medical LLM. This superiority in predictive accuracy translates directly into a greater concordance with established pharmacovigilance signals and the identification of more clinically nuanced causal factors, supporting the hypothesis that a better predictive model is also a better tool for causal discovery. Our work underscores the paramount importance of domain specialization over raw scale in building effective AI for healthcare and provides a clear, evidence-based directive for future efforts in computational pharmacovigilance: prioritize deep, relevant domain knowledge in model selection.

\section*{Declaration of AI assistance}

The author(s) used Gemini 3 Pro for language editing and polishing. After using this technology, the author(s) reviewed the results and take(s) full responsibility for the contents of the manuscript.

\section*{Declaration of funding}

The work of Roland Molontay has been supported by the János Bolyai Research Scholarship of the Hungarian Academy of Sciences.
Csaba Kiss has been supported by the Cooperative Doctoral Programme of the National Research, Development, and Innovation Fund.

\bibliographystyle{plain}
\bibliography{bibliography}

\end{document}